\documentclass[10pt,twocolumn,letterpaper]{article}

\usepackage{cvpr}
\usepackage{times}
\usepackage{epsfig}
\usepackage{graphicx}
\usepackage{amsmath}
\usepackage{amssymb}
\usepackage{booktabs}
\usepackage{caption}
\usepackage{subcaption}
\usepackage{multirow}
\usepackage{bm}
\usepackage{balance}
\usepackage{enumitem}
\usepackage[accsupp]{axessibility}
\graphicspath{ {./images/} }

\usepackage[pagebackref,breaklinks,colorlinks]{hyperref}

\usepackage[symbol]{footmisc}

\renewcommand{\thefootnote}{\fnsymbol{footnote}}

\begin{document}

\title{Identity-Preserving Aging of Face Images via Latent Diffusion Models}

\author{Sudipta Banerjee\thanks{Both authors contributed equally.}
\and
Govind Mittal\footnotemark[1]
\and
Ameya Joshi
\and
Chinmay Hegde
\and
Nasir Memon \\
New York University\\
{\tt\small \{sb9084, mittal, ameya.joshi, chinmay.h, memon\}@nyu.edu}
}

\maketitle
\thispagestyle{empty}
\renewcommand*{\thefootnote}{\arabic{footnote}}
\setcounter{footnote}{0}
\begin{abstract}
 The performance of automated face recognition systems is inevitably impacted by the facial aging process. However, high quality datasets of individuals collected over several years are typically small in scale. In this work, we propose, train, and validate the use of latent text-to-image diffusion models for synthetically aging and de-aging face images. Our models succeed with few-shot training, and have the added benefit of being controllable via intuitive textual prompting. We observe high degrees of visual realism in the generated images while maintaining biometric fidelity measured by commonly used metrics. We evaluate our method on two benchmark datasets (CelebA and AgeDB) and observe significant reduction ($\sim44\%$) in the False Non-Match Rate compared to existing state-of the-art baselines.

\end{abstract}

\section{Introduction}
\label{sec:intro}

\noindent\textbf{Motivation.} It is well known that facial aging can significantly degrade the performance of modern automated face recognition systems ~\cite{NIST2,Introtobiom,NIST1}. Improving the robustness of such systems to aging variations is therefore critical for their lasting practical use. However, building systems that are robust to aging variations requires high quality longitudinal datasets: images of a large number of individuals collected over several years. Collection of such data constitutes a major challenge in practice. Datasets such as MORPH~\cite{MORPH} contains longitudinal samples of only 317 subjects from a total of $\sim$13K subjects over a period of five years~\cite{Lacy_Jain_2018}. Other datasets like AgeDB~\cite{agedb} and CACD~\cite{CACD} contains unconstrained images with significant variations in pose, illumination, background, and expression. 

An alternative approach to gathering longitudinal data is to digitally simulate face age progression~\cite{dcface}.
Approaches include manual age-editing tools, such as YouCam Makeup, FaceApp, and AgingBooth~\cite{ageapp,FaceAPP}; more recently, GAN-based generative models, such as AttGAN, Cafe-GAN, Talk-to-Edit~\cite{AttGAN, talktoedit, CafeGAN, STGAN, GANpyramid} have also been used to simulate age progression in face images. However, we find that generative models struggle to correctly model biological aging, which is a complex process affected by genetic, demographic, and environmental factors. 
Moreover, training high quality GANs for adjusting facial attributes themselves require a large amount of training data. 

\begin{figure*}[t]
     \centering
     
         \includegraphics[width=0.85\textwidth]{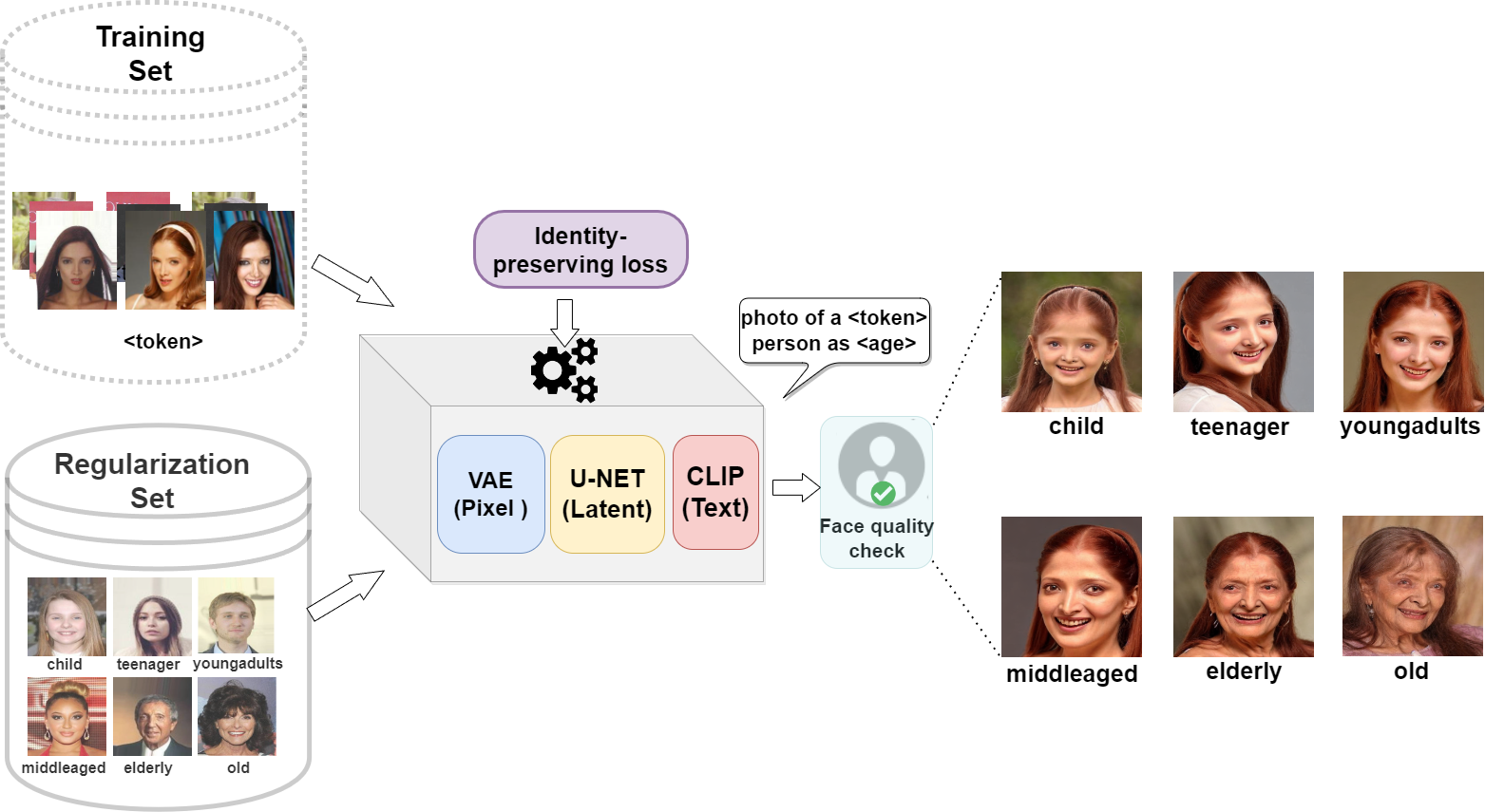}
         
        \caption{Overview of the proposed method. The proposed method needs a fixed \textit{Regularization Set} comprising facial images with age variations and a variable \textit{Training Set} comprising facial images of a target individual. The latent diffusion module (comprising a VAE, U-Net and CLIP-text encoder) learns the concept of age progression from the regularization images and the identity-specific information from the training images. We integrate biometric and contrastive losses in the network for identity preservation. At inference, the user prompts the trained model using a rare token associated with the trained target subject and the desired age to perform age editing. }
        \label{fig:overview}
\end{figure*}

\noindent\textbf{Our Approach.} Existing generative models often struggle to manipulate the age attribute and preserve facial identity. They also require auxiliary age classifiers and/or extensive training data with longitudinal age variations. To address both of the above issues, we propose a new latent generative model for simulating high-quality facial aging, while simultaneously preserving biometric identity. The high level algorithmic idea is to finetune latent text-to-image diffusion models (such as Stable Diffusion~\cite{Latentdiff}) with a novel combination of contrastive and biometric losses that help preserve facial identity. 
See Fig.~\ref{fig:overview} for an overview of our method. 

The proposed method requires: (i) a pre-trained latent diffusion model (see Sec.~\ref{sec:rel}), (ii) a small set (numbering $\approx$ 20) of training face images of an individual, and (iii) a small auxiliary set (numbering $\approx$ 600) of image-caption pairs. The pairs contain facial images of individuals and captions indicating their corresponding age. This auxiliary set of image-caption pairs serve as the regularization set. The individuals in the training set and the regularization set are disjoint. We use the training images during fine-tuning to learn the identity-specific information of the individual, and the regularization images with captions to learn the association between an image (face) and its caption (age). Finally, we simulate age regression and progression of the trained individual using a text prompt specifying the target age. See the details of our method in Sec.~\ref{sec:prop}. 

\noindent\textbf{Summary}. Our main contributions are as follows. 
\begin{itemize}[leftmargin=*]
    \item We adapt latent diffusion models to perform age regression and progression in face images. We introduce two key ideas: an identity-preserving loss (in addition to perceptual loss), and a small regularization set of image-caption pairs to resolve the limitations posed by existing GAN-based methods. 
    \item As a secondary finding, we show that face recognition classifiers may benefit by fine-tuning on generated images with significant age variations as indicated in~\cite{Synface}. 
    \item We conduct experiments on CelebA and AgeDB datasets and perform evaluations to demonstrate that the synthesized images i) appear visually compelling in terms of aging and de-aging through qualitative analysis and automated age predictor, and ii) match with the original subject with respect to human evaluators and automated face matcher. We demonstrate that our method outperforms SOTA image editing methods, namely, IPCGAN~\cite{IPCGAN}, AttGAN~\cite{AttGAN} and Talk-to-Edit~\cite{talktoedit}.
\end{itemize} 

The rest of the paper is organized as follows. Sec.~\ref{sec:rel} outlines existing work. Sec.~\ref{sec:prop} describes the proposed method for simulating facial aging and de-aging. Sec.~\ref{sec:exp} describes the experimental settings. Sec.~\ref{sec:find} presents our findings and analysis. Sec.~\ref{sec:sum} concludes the paper.

\section{Related Work}
\label{sec:rel}

Previous automated age progression models have used a variety of architectures, including recurrent ones~\cite{RecurrentFA} and GANs. 
 \cite{GANpyramid} uses a hierarchy of discriminators to preserve the reconstruction details, age and identity. STGAN~\cite{STGAN} utilizes selective transfer units that accepts the difference between the target and source attribute vector as input, resulting in more controlled manipulation of the attribute. Cafe-GAN~\cite{CafeGAN} utilizes complementary attention features to focus on the regions pertinent to the target attribute while preserving the remaining details. 
HRFAE~\cite{hrfae} encodes an input image to a set of age-invariant features and an age-specific modulation vector. The age-specific modulation vector re-weights the encoded features depending on the target age and then passes it to a decoder unit that edits the image. CUSP~\cite{cusp} uses a custom structure preserving module that masks the irrelevant regions for better facial structure preservation in the generated images. The method performs style and content disentanglement while conditioning the generated image on the target age. ChildGAN~\cite{ChildGAN} is inspired from the self-attention GAN and uses one-hot encoding of age labels and gender labels appended to the noise vector to perform age translation in images of young children. 

We focus on three methods in our comparisons. IPCGAN~\cite{IPCGAN} uses a conditional GAN with an identity preserving module and an age classifier to perform image-to-image style transfer for age-editing. AttGAN~\cite{AttGAN} performs binary facial attribute manipulation by modeling the relationship between the attributes and the latent representation of the face. The network enables high quality facial attribute editing while controlling the attribute intensity and style.
Talk-to-Edit~\cite{talktoedit} provides fine-grained facial attribute editing via dialog interaction, similar to our approach. The method uses a language encoder to convert the user's request into an `editing encoding' that encapsulates information about the degree and direction of change of the target attribute, and seeks user feedback to iteratively edit the desired attribute. 

We also highlight two recent methods that also use diffusion models for face generation. In DCFace~\cite{dcface}, the authors propose a dual condition synthetic face generator to allow control over simulating intra-class (within same individual) and inter-class (across different individuals) variations. In~\cite{Amerini}, the authors explore suitable prompts for generating realistic faces using stable diffusion and investigate their quality. Neither method focus on identity-preserving text guided facial aging and de-aging, which is our goal.

\section{Our Proposed Method}
\label{sec:prop}
Although a suite of age editing methods exist in the literature as discussed above, the majority of them focuses on perceptual quality instead of biometric quality. A subset of latent space manipulation methods struggle with `real' face images and generate unrealistic outputs. Existing works reiterate that age progression is a smooth but non-deterministic process that requires incremental evolution to effectively transition between ages. This motivates the use of diffusion models, which naturally model the underlying data distribution by incrementally adding and removing noise. We start with a brief mathematical overview.

\subsection{Preliminaries}
Denoising diffusion probabilistic models (DDPMs)~\cite{DDPM} perform the following steps: 1) a forward diffusion process $\bm{x}_0 \xrightarrow{>> \eta_t} \bm{x}_t$ \footnote{\label{note1}$>>$ denotes noise addition while $<<$ denotes noise removal.} that incrementally adds Gaussian noise, $\eta$ sampled from a normal distribution, $\mathcal{N} (\bm{0},\bm{I})$, to the clean data, $\bm{x}_0$ sampled from a real distribution, $p(\bm{x})$ over $t$ time steps. 2) a backward denoising process $\bm{x}_0 \xleftarrow{<< \eta_t} \bm{x}_t$ \footref{note1} that attempts to recover the clean data from the corrupted or noisy data $\bm{x}_t$ by approximating the conditional probability distribution, $p(\bm{x}_{t-1} \mid \bm{x}_t)$ using a neural network that serves as a noise estimator. The forward and backward processes can be considered analogous to VAEs~\cite{VAE}.

Note that DDPMs are computationally expensive as the estimated noise has the same dimension as the input. Alternatively, stable diffusion~\cite{Latentdiff} is a class of latent diffusion models that performs diffusion on a relatively lower dimensional latent representation. Latent diffusion generates high quality images conditioned on text prompts. It comprises three modules: an autoencoder (VAE), a U-Net and a text-encoder. The encoder in the VAE converts the image into a low dimensional latent representation fed as the input to the U-Net model. The U-Net model estimates the noise needed to recover the high resolution output from the decoder of the VAE. \cite{Latentdiff} further added cross-attention layers in the U-Net backbone to use text embedding as a conditional input, thereby enhancing the model's generative capability. 

In this work, we focus on DreamBooth~\cite{dreambooth}, a latent diffusion model that fine-tunes a text-to-image diffusion framework for re-contextualization of a single subject. To accomplish this, it requires (i) a few images of the subject, and (ii) text prompts containing a unique identifier and the class label of the subject. The class label denotes a collective representation of multiple instances while the subject will correspond to a specific example belonging to the class. The objective is to associate a unique token or a rare identifier to each subject (a specific instance of a class) and then recreate images of the same subject in different contexts as guided by the text prompts. The class label harnesses the prior knowledge of the trained diffusion framework for that specific class. Incorrect class labels or missing class labels may result in inferior outputs~\cite{dreambooth}. The unique token acts as a reference to the particular subject, and needs to be rare enough to avoid conflict with other concepts. The authors use a set of rare tokens corresponding to a sequence of 3 or fewer Unicode characters and the T5-XXL tokenizer. See~\cite{dreambooth} for more details. DreamBooth uses a class-specific prior preservation loss to increase the variability of generated images while ensuring minimal deviation between the target subject and the output images. The original training loss can be written as follows.
\begin{equation}
\label{Eqn:SD}
\begin{split}
    \mathbb{E}_{\bm{x}, \bm{c}, t}[w_t &\|f_{\theta}(g_t(\bm{x}),c) - \bm{x}\|_2^2 + \\ & \lambda w_{t'} \| f_{\theta}(g_{t'}(\bm{x'}),c_{class}) - \bm{x'}\|_2^2].
    \end{split}
\end{equation}
The first term in Eqn.~\ref{Eqn:SD} denotes the squared error between the ground-truth images, $\bm{x}$, (training set) and the generated images, $f_{\theta}(g_t(\bm{x}),c)$. Here, $f_{\theta}(\cdot , \cdot)$ denotes the pre-trained diffusion model (parameterized by $\theta$) that generates images for a noise map and a conditioning vector. The noise map is obtained as $g_t(\bm{x}) = \alpha_t\bm{x} + \sigma_t \eta$, where $\eta \sim \mathcal{N} (\bm{0}, \bm{I})$, and $\alpha_t, \sigma_t, w_t$ are diffusion control parameters at time step $t \sim\mathbb{U}[0,1]$. The conditioning vector $\bm{c}$ is generated using a text encoder for a user-defined prompt. 
  The second term refers to the prior-preservation component using generated images that represents the prior knowledge of the trained model for the specific class. The term is weighted by a scalar value, $\lambda = 1$. The conditioning vector in the second term, $\bm{c}_{class}$, corresponds to the class label. 

\subsection{Methodology}
DreamBooth works effectively with the aid of prior preservation for synthesizing images of dogs, cats, cartoons, etc. But in this work, we are focusing on human face images that contain intricate structural and textural details. Although the class label `person' can capture human-like features, this may not be adequate to capture identity-specific features that vary across individuals. Therefore, we include an identity-preserving term in the loss function. The identity-preserving component minimizes the distance between the biometric features from the original and generated images as follows. 
\begin{equation}
\label{Eqn:SD_biom}
\begin{split}
 \mathbb{E} & _{\bm{x}, \bm{c}, t}[w_t \|f_{\theta}(g_t(\bm{x}),c) - \bm{x}\|_2^2 + \\ & \lambda w_{t'} \| f_{\theta}(g_{t'}(\bm{x'}),c_{class}) - \bm{x'}\|_2^2 + \\& \lambda_{b}\mathcal{B}( f_{\theta}(g_{t}(\bm{x}),c_{class}), \bm{x})].
    \end{split}
\end{equation}

We use this new loss to fine-tune the VAE. The third term in Eqn.~\ref{Eqn:SD_biom} refers to the biometric loss computed between the ground-truth image of the subject, $\bm{x}$, and the generated image weighted by $\lambda_{b} = 0.1$. Note that $f_{\theta}(g_{t'}(\bm{x}),c_{class})$ uses the training set (\textit{i.e.}, images of an individual subject), whereas $f_{\theta}(g_{t'}(\bm{x'}),c_{class})$ uses the regularization set that contains representative images of a class. Here, $\mathcal{B}(\cdot , \cdot)$ computes the $L_1$ distance between the biometric features extracted from a pair of images (close to zero for same subjects, higher values correspond to different subjects). We use a pre-trained VGGFace~\cite{facenet} feature extractor, such that, $$\mathcal{B}(i , j) = \left\| VGGFace(i) - VGGFace(j)\right\|_1 .$$



Now, we turn to target-specific fine-tuning. The implementation used in our work~\cite{imp, textualinv} uses a frozen VAE and a text-encoder while keeping the U-Net model unfrozen. U-Net denoises the latent representation produced by the encoder of VAE, $g_t(\bm{x}) = \bm{z}_t = \alpha_t \bm{x} + \sigma_t \eta$. Therefore, we use identity-preserving contrastive loss using the latent representation. We adopted the SimCLR~\cite{SimCLR} framework that uses a normalized temperature-scaled cross-entropy loss between positive and negative pairs of augmented latent representations, denoted by $\mathcal{S}(\cdot , \cdot)$ in Eqn.~\ref{Eqn:SD_contrast}. We compute the contrastive loss between the latent representation of the noise-free inputs ($\bm{z}_0$) and the de-noised outputs ($\bm{z}_t$) with a weight term $\lambda_s = 0.1$ and a temperature value = 0.5. Refer to~\cite{SimCLR} for more details. The contrastive loss between the latent representation in the U-Net architecture enables us to fine-tune the diffusion model for each subject as follows. 

\begin{equation}
\label{Eqn:SD_contrast}
\begin{split}
  \mathbb{E} & _{\bm{x}, \bm{c}, t}[w_t \|f_{\theta}(g_t(\bm{x}),c) - \bm{x}\|_2^2 + \\ & \lambda w_{t'} \| f_{\theta}(g_{t'}(\bm{x'}),c_{class}) - \bm{x'}\|_2^2 + \lambda_{s} \mathcal{S} (\bm{z}_{t} , \bm{z}_{0})].
    \end{split}
\end{equation}


In addition to customizing the losses, we use the regularization set to impart the concept of facial age progression and regression to the latent diffusion model. The regularization set contains representative images of a class, in our case, `person'. A regularization set comprising face images selected from the internet would have sufficed if our goal was to generate realistic faces as done in~\cite{Amerini}. However, our task involves learning the concept of aging and de-aging, and then apply it to any individual. To accomplish this task, we use face images from different age groups and then pair it with one-word captions that indicate the age group of the person depicted in the image. The captions correspond to one of the six age groups: 
 `child', `teenager', `youngadults', `middleaged', `elderly', and `old'. We could have used numbers as age groups, for example, twenties, forties or sixties, but we found that a language description is more suitable than a numeric identifier. Another reason for pairing these age descriptions with the images is that we can use these same age identifiers while prompting the diffusion model during inference (photo of a $\langle$ token $\rangle$ $\langle$ class label $\rangle$ as $\langle$ age group $\rangle$). We use the following six prompts during inference. 1) photo of a sks person as child, 2) photo of a sks person as teenager, 3) photo of a sks person as youngadults, 4) photo of a sks person as middleaged, 5) photo of a sks person as elderly, and 6) photo of a sks person as old. We have explored other tokens (see Sec.~\ref{token}).

\begin{figure*}[t]
     \centering
     \begin{subfigure}[b]{0.12\textwidth}
         \centering
         \caption*{\textbf{original}}
         \includegraphics[width=\textwidth, height=\textwidth]{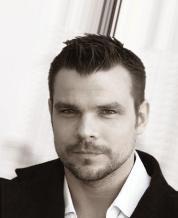}  
     \end{subfigure}
     \begin{subfigure}[b]{0.12\textwidth}
         \centering
         \caption*{child}
         \includegraphics[width=\textwidth]{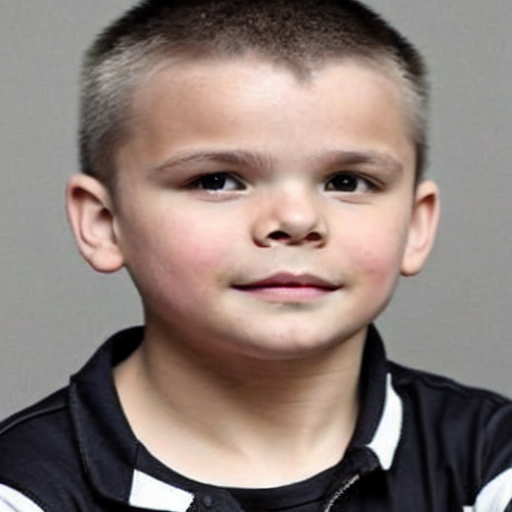}  
     \end{subfigure} 
     \begin{subfigure}[b]{0.12\textwidth}
         \centering
         \caption*{teenager}
         \includegraphics[width=\textwidth]{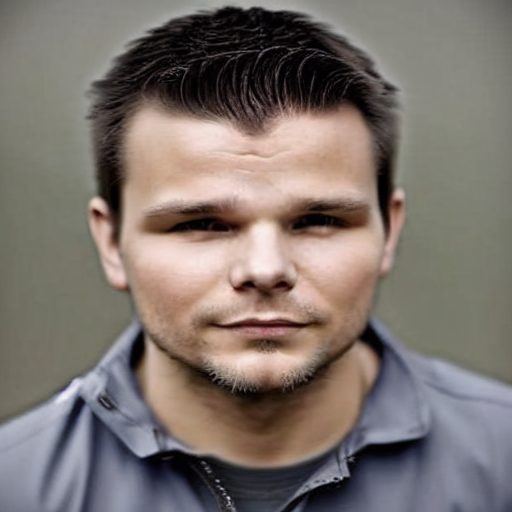}
     \end{subfigure}
     \begin{subfigure}[b]{0.12\textwidth}
         \centering
         \caption*{youngadults}
         \includegraphics[width=\textwidth]{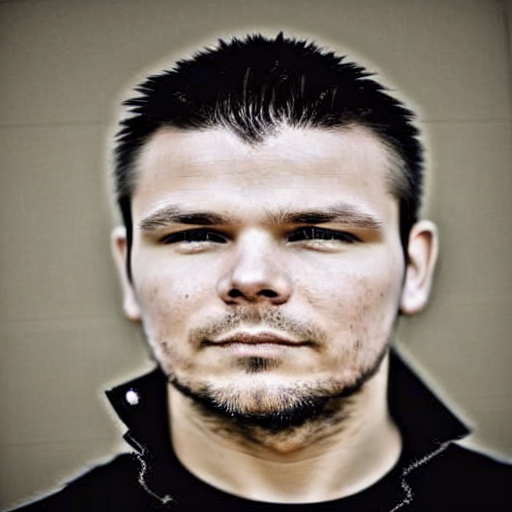}
     \end{subfigure}
     \begin{subfigure}[b]{0.12\textwidth}
         \centering
         \caption*{middleaged}
         \includegraphics[width=\textwidth]{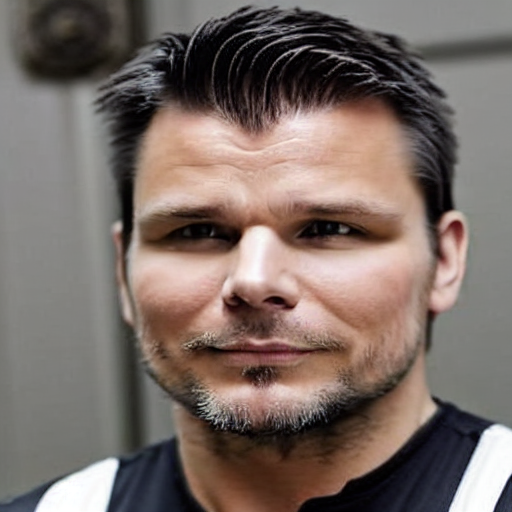}
     \end{subfigure}
     \begin{subfigure}[b]{0.12\textwidth}
         \centering
         \caption*{elderly}
         \includegraphics[width=\textwidth]{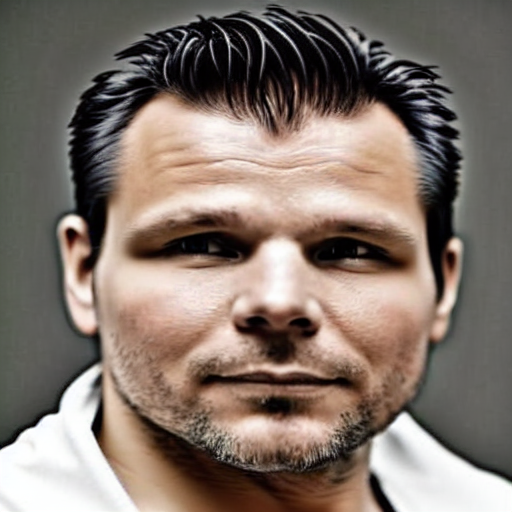}
     \end{subfigure}
     \begin{subfigure}[b]{0.12\textwidth}
         \centering
         \caption*{old}
         \includegraphics[width=\textwidth]{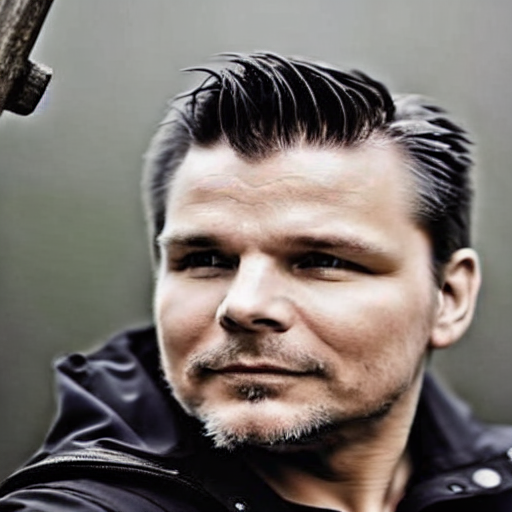}
     \end{subfigure} \\
     \begin{subfigure}[b]{0.12\textwidth}
         \centering
         \includegraphics[width=\textwidth, height=\textwidth]{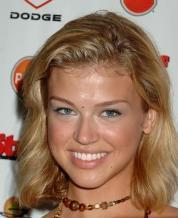}  
     \end{subfigure}
     \begin{subfigure}[b]{0.12\textwidth}
         \centering
         \includegraphics[width=\textwidth]{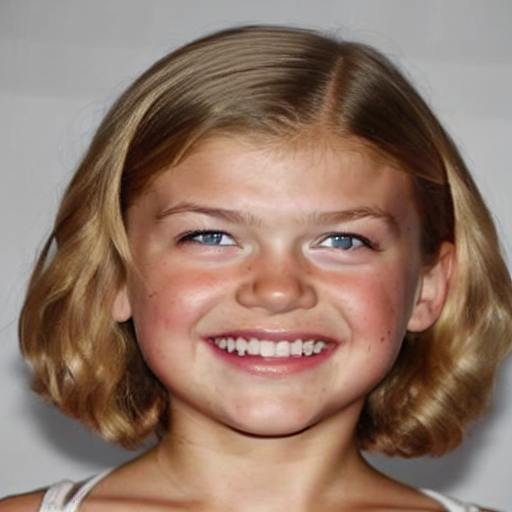}
     \end{subfigure} 
     \begin{subfigure}[b]{0.12\textwidth}
         \centering
         \includegraphics[width=\textwidth]{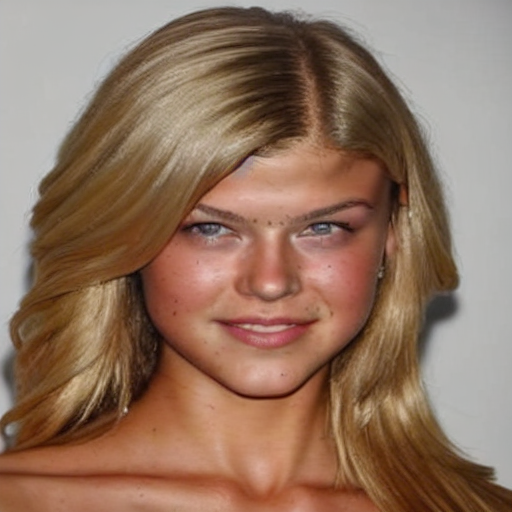}
     \end{subfigure}
     \begin{subfigure}[b]{0.12\textwidth}
         \centering
         \includegraphics[width=\textwidth]{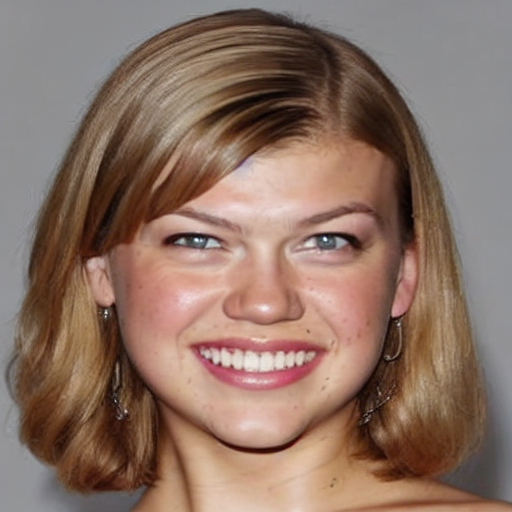}
     \end{subfigure}
     \begin{subfigure}[b]{0.12\textwidth}
         \centering
         \includegraphics[width=\textwidth]{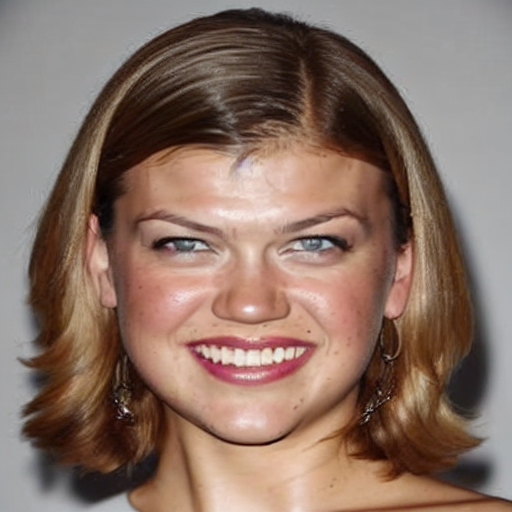}
     \end{subfigure}
     \begin{subfigure}[b]{0.12\textwidth}
         \centering
         \includegraphics[width=\textwidth]{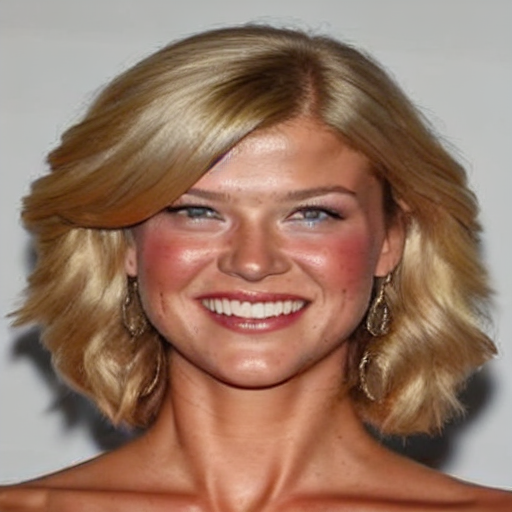}
     \end{subfigure}
     \begin{subfigure}[b]{0.12\textwidth}
         \centering
         \includegraphics[width=\textwidth]{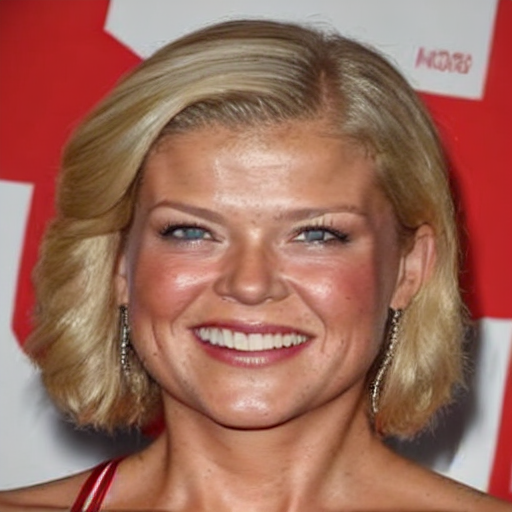}
     \end{subfigure} 
        \caption{Illustration of age edited images generated from the CelebA dataset.}
        \label{fig:Examples_CelebA}
\end{figure*}

\section{Experiments}
\label{sec:exp}

\textbf{Setup and implementation details. }
We conduct experiments using DreamBooth implemented using Stable Diffusion v1.4 ~\cite{imp}. The model uses CLIP's~\cite{clip} text encoder trained on laion-aesthetics v2 5+ and a vector quantized VAE~\cite{vqvae} to accomplish the task of age progression. The text encoder stays frozen while training the diffusion model. We use two datasets, namely, \textbf{CelebA}~\cite{CelebA} and \textbf{AgeDB}~\cite{agedb}. We use 2,258 face images belonging to 100 subjects from the CelebA~\cite{CelebA} dataset, and 659 images belonging to 100 subjects from the AgeDB dataset to form the `training set'. CelebA does not contain age information, except a binary `Young' attribute annotation. We do not have ground-truth for evaluating the generated images synthesized from the CelebA dataset. On the other hand, AgeDB dataset comprises images with exact age values. We then select the age group that has the highest number of images and use them as the training set, while the remaining images contribute towards the testing set. Therefore, 2,369 images serve as ground-truth for evaluation in AgeDB dataset. 

 We use a regularization set comprising image-caption pairs where each face image is associated with a caption indicating its corresponding age label. We use 612 images belonging to 375 subjects from the CelebA-Dialog~\cite{talktoedit} dataset,where the authors provide fine-grained annotations of age distributions. We convert the distribution to categorical labels to use as captions in the regularization images. We refer to them as \{Child: $<$15 years, Teenager: 15-30 years, Youngadults: 30-40 years, Middleaged: 40-50 years, Elderly: 50-65 years and Old: $>$65 years\}. We use 612 ($102\times6$) images in the subject disjoint regularization set.

The success of generating high quality images often depend on effectively prompting the diffusion model during inference. The text prompt at the time of inference needs a rare token/identifier that is associated with the concept learnt during fine-tuning. We use four different rare tokens \{\textit{wzx}, \textit{sks}, \textit{ams}, \textit{ukj}\}~\cite{reddit} in this work for brevity.

We use the implementation of DreamBooth using stable diffusion in~\cite{imp} and used the following hyperparameters. We adopt a learning rate = 1e-6, number of training steps $=800$, embedding dimensionality in autoencoder $=4$, and batch size $=8$. The generated images are of size $512 \times 512$. We use $\lambda=1, \lambda_b=0.1$ and $\lambda_s=0.1$ (refer to Eqns.~\ref{Eqn:SD_biom} and ~\ref{Eqn:SD_contrast}). We generate 8 samples at inference. However, we perform a facial quality assessment using EQFace~\cite{EQFace} to limit the number of generated face images to 4, such that, each generated image contains a single face with frontal pose. We adopt a threshold of 0.4, and retain the generated images if quality exceeds the threshold, else, discard them. Training each subject requires $\sim$5-8 mins. on a A100 GPU.

\begin{table}[b]
\centering
\caption{CelebA simulation results for biometric matching between \textbf{Original-Modified} images. The metrics are False Non-Match Rate (FNMR) at False Match Rate (FMR) = 0.01/0.1\%.}
\label{Tab:Sim}
\scalebox{0.92}{
\begin{tabular}{l||cc||cc} \hline
\multirow{2}{*}{\textbf{Age group}} & \multicolumn{2}{c||}{\textbf{Initial loss}} & \multicolumn{2}{c}{\textbf{With contrastive loss}} \\
                           & sks          & wzx          & sks          & wzx         \\\hline \hline
child                      &  0.49/0.21            &  0.58/0.27            &   0.56/0.26           & 0.60/0.29            \\
teenager                   &  0.23/0.07              & 0.32/0.12             &  0.29/0.10            & 0.34/0.12            \\
youngadults                &  0.25/0.08          &  0.30/0.10            &   0.28/0.08           & 0.31/0.10            \\
middleaged                 & 0.20/0.07           &  0.28/0.09            &   0.27/0.09           &   0.30/0.10          \\
elderly                    &  0.22/0.07            & 0.29/0.10             &    0.25/0.09          &  0.29/0.11           \\
old                        &   0.24/0.10           & 0.31/0.12            &   0.29/0.11           &  0.32/0.12          \\ \hline
\end{tabular}}
\end{table}

\begin{figure}[b]
     \centering
        \begin{subfigure}[t]{0.92\columnwidth}
        \centering
         \includegraphics[width=\columnwidth]{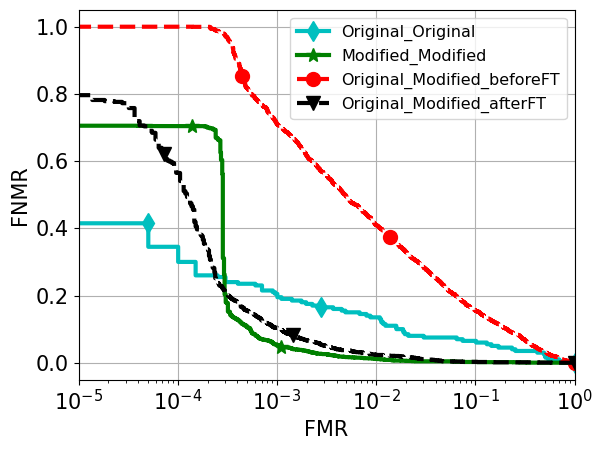}
        \end{subfigure}

    \begin{subfigure}[h]{0.92\textwidth}
    \begin{tabular}{l||l} \hline
    \textbf{Matching scenarios } & \textbf{FNMR@FMR=0.01/0.1\%} \\ \hline \hline
    Ori-Ori   & 0.14/0.07      \\
    Mod-Mod   & 0.02/0.01         \\
    Ori-Mod (w/o fine-tune) & 0.41/0.16         \\
    Ori-Mod (w/ fine-tune)       & \textbf{0.03/0.01 } \\ \hline      
    \end{tabular}
    \end{subfigure}
    \vspace{0.5em}
    \caption{(Top:) DET curves of face matching using generated images from the CelebA dataset. (Bottom:) Recognition performance in the table indicating FNMR @ FMR=0.01/0.1\%. The age-edited images are generated using the \textit{wzx} token with contrastive loss. }
    \label{Fig:CelebA_Mod}
\end{figure}

\begin{figure*}[t]
     \centering
     \begin{subfigure}[b]{0.12\textwidth}
         \centering
         \caption*{\textbf{original}}
         \includegraphics[width=\textwidth]{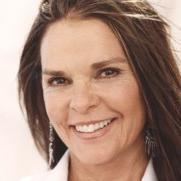}  
     \end{subfigure} 
     \begin{subfigure}[b]{0.12\textwidth}
         \centering
         \caption*{child}
         \includegraphics[width=\textwidth]{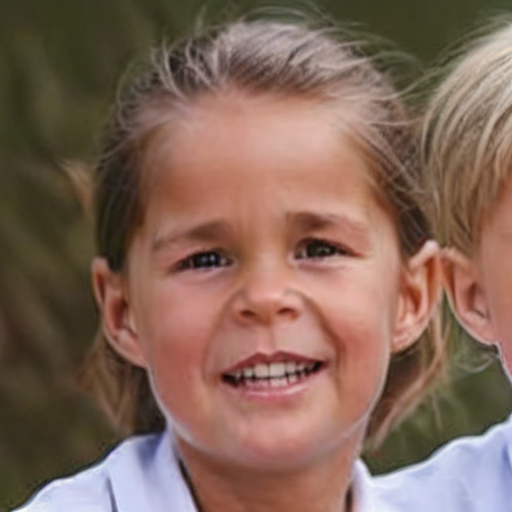}  
     \end{subfigure} 
     \begin{subfigure}[b]{0.12\textwidth}
         \centering
         \caption*{teenager}
         \includegraphics[width=\textwidth]{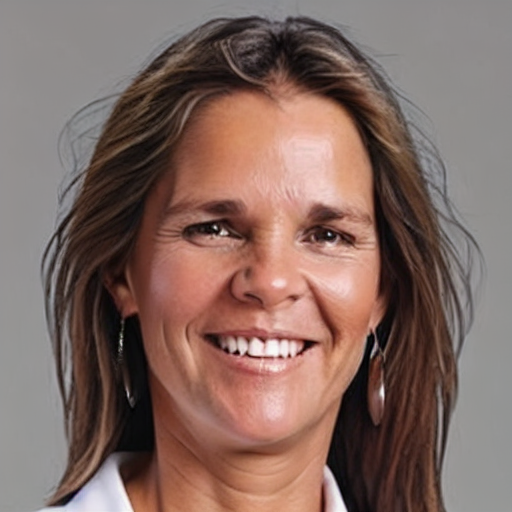}
     \end{subfigure}
     \begin{subfigure}[b]{0.12\textwidth}
         \centering
         \caption*{youngadults}
         \includegraphics[width=\textwidth]{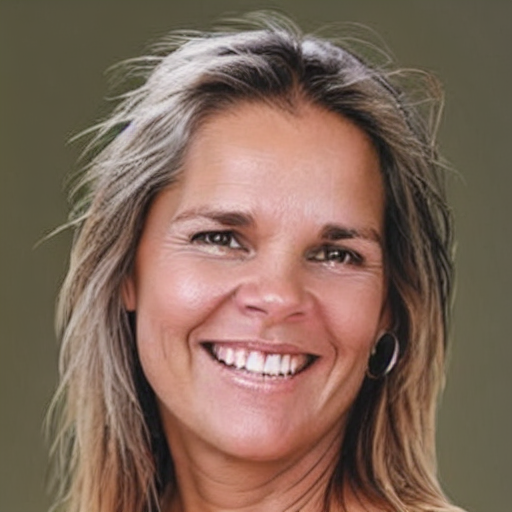}
     \end{subfigure}
     \begin{subfigure}[b]{0.12\textwidth}
         \centering
         \caption*{middleaged}
         \includegraphics[width=\textwidth]{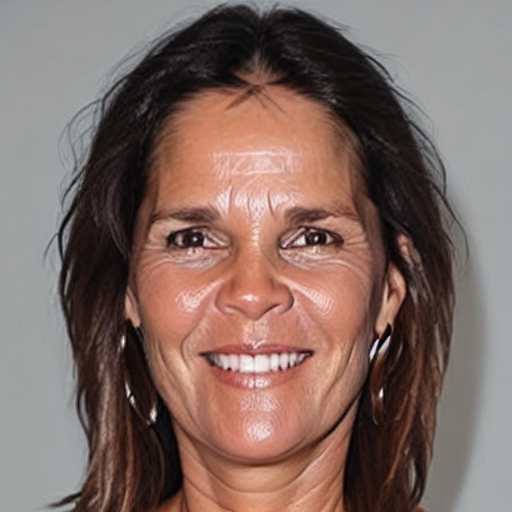}
     \end{subfigure}
     \begin{subfigure}[b]{0.12\textwidth}
         \centering
         \caption*{elderly}
         \includegraphics[width=\textwidth]{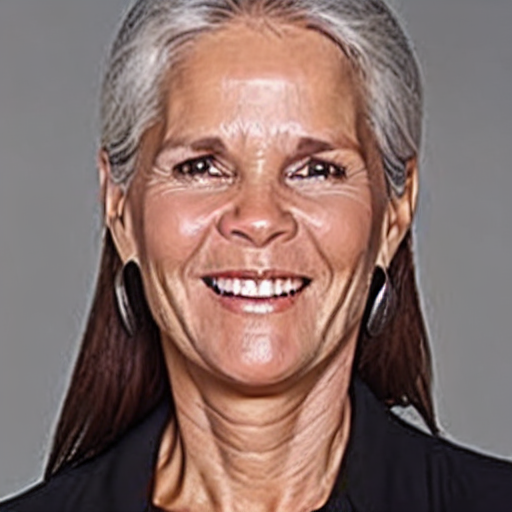}
     \end{subfigure}
     \begin{subfigure}[b]{0.12\textwidth}
         \centering
         \caption*{old}
         \includegraphics[width=\textwidth]{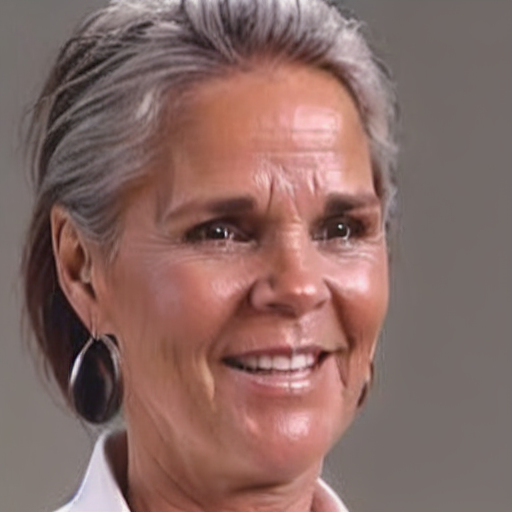}
     \end{subfigure} \\
     \begin{subfigure}[b]{0.12\textwidth}
         \centering
         \includegraphics[width=\textwidth]{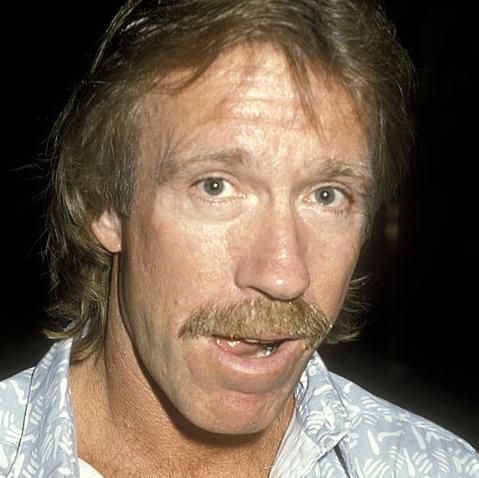}  
     \end{subfigure} 
     \begin{subfigure}[b]{0.12\textwidth}
         \centering
         \includegraphics[width=\textwidth]{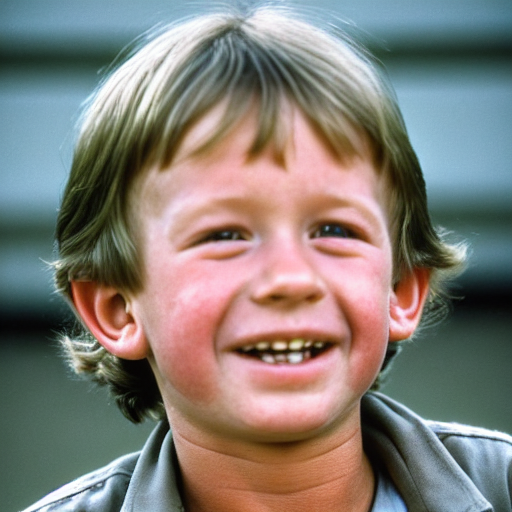}  
     \end{subfigure} 
     \begin{subfigure}[b]{0.12\textwidth}
         \centering
         \includegraphics[width=\textwidth]{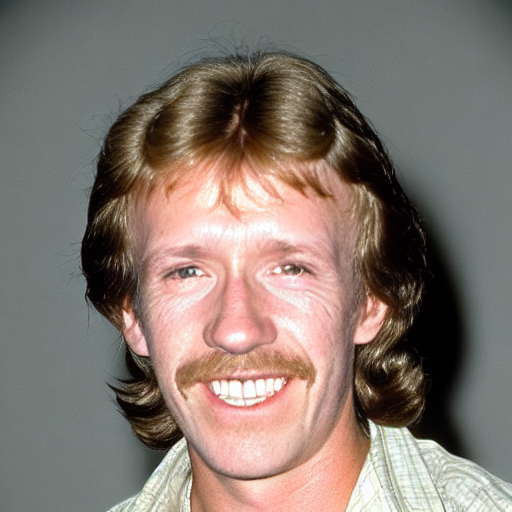}
     \end{subfigure}
     \begin{subfigure}[b]{0.12\textwidth}
         \centering
         \includegraphics[width=\textwidth]{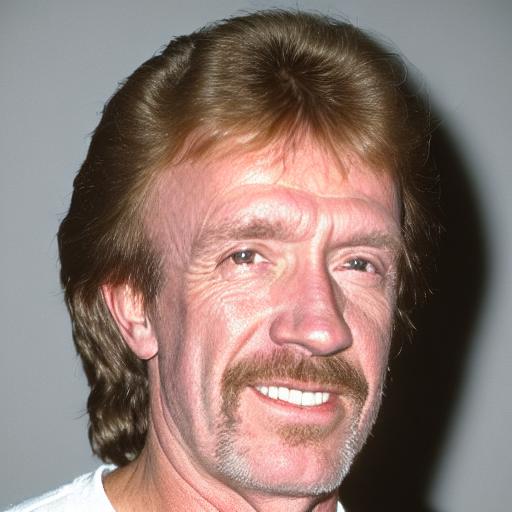}
     \end{subfigure}
     \begin{subfigure}[b]{0.12\textwidth}
         \centering
         \includegraphics[width=\textwidth]{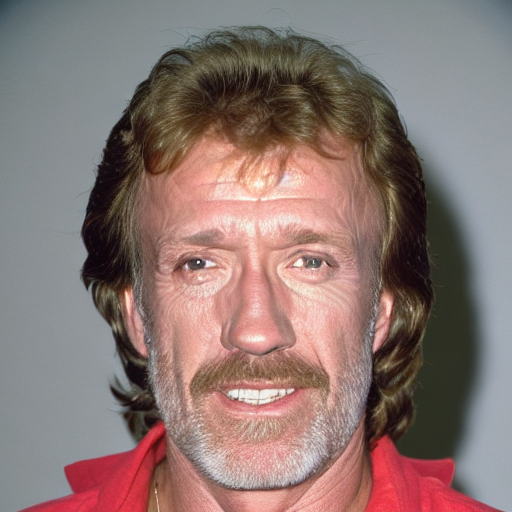}
     \end{subfigure}
     \begin{subfigure}[b]{0.12\textwidth}
         \centering
         \includegraphics[width=\textwidth]{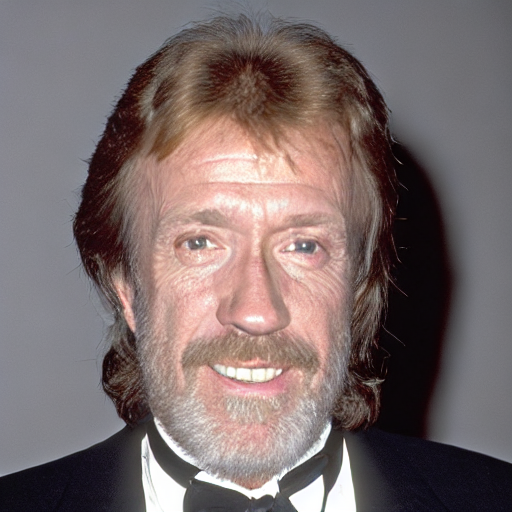}
     \end{subfigure}
     \begin{subfigure}[b]{0.12\textwidth}
         \centering
         \includegraphics[width=\textwidth]{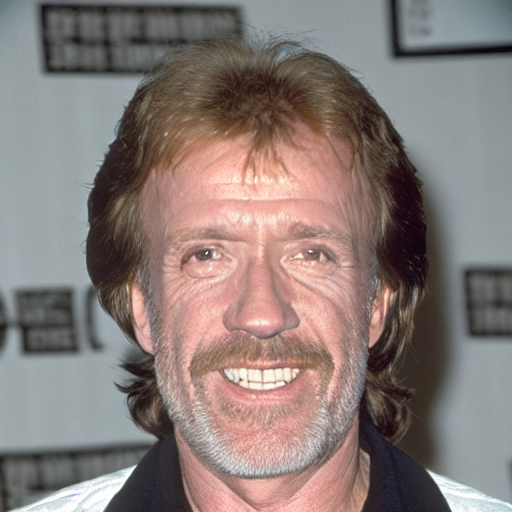}
     \end{subfigure} 
     \caption{Illustration of age edited images generated from the AgeDB dataset.}
        \label{fig:Examples_ageDB}
\end{figure*}

\begin{figure}[ht!]
     \centering
        
        \begin{subfigure}[h]{0.92\columnwidth}
        \centering
         \includegraphics[width=\columnwidth]{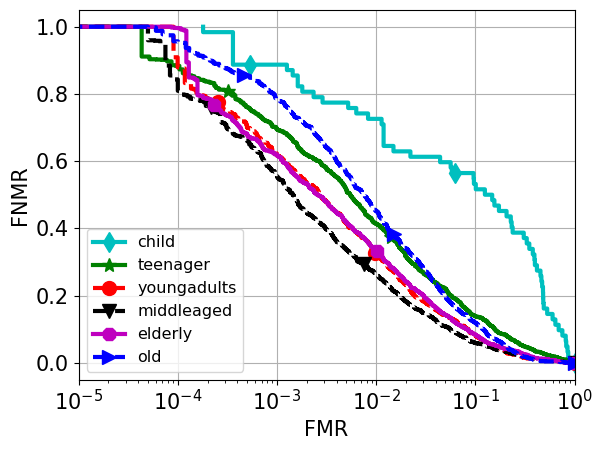}
        \end{subfigure}

    \begin{subfigure}[h]{0.92\textwidth}
    \hspace{3.0em}
    \begin{tabular}{l||l} \hline
    \textbf{Age group} & \textbf{FNMR@FMR=0.01/0.1\%} \\ \hline \hline
    child   & 0.73/0.54      \\
    teenager   & 0.42/0.14        \\
    youngadults & 0.33/0.09         \\
    middleaged & 0.27/0.06 \\ 
    elderly & 0.34/0.09 \\
    old &  0.46/0.12 \\ \hline
    \end{tabular}
    \end{subfigure}
    \vspace{0.5em}
    \caption{(Top:) DET curves of face matching using generated images from the AgeDB dataset for the six age groups. (Bottom:) Recognition performance in the table indicating FNMR @ FMR=0.01/0.1\%. The age-edited images are generated using the \textit{wzx} token with contrastive loss. }
    \label{fig:agedbDET}
\end{figure}


We perform \textbf{qualitative evaluation} of the generated images by conducting a user study involving 26 volunteers. The volunteers are shown a set of 10 face images (original) and then 10 generated sets; each set contains five images belonging to five age groups (excluding old), resulting in a total of 60 images. They are assigned two tasks: 1) identify the individual from the original set who appears most similar to the subject in the generated set; 2) assign each of the five generated images to the five age groups they are most likely to belong to. We compute the proportion of correct face recognition and age group assessment.

 Further, we perform \textbf{quantitative evaluation} of the generated outputs using the ArcFace~\cite{ArcFace} matcher (different from VGGFace used in identity-preserving biometric loss). We utilize the genuine (intra-class) and imposter (inter-class) scores to compute Detection Error Trade-off (DET) curves and report the False Non-Match Rate (FNMR) at a False Match Rate (FMR) of 0.01\% and 0.1\%.

\section{Results}
\label{sec:find}

We report the biometric matching performance using the ArcFace matcher between \textbf{original and modified} images in Table~\ref{Tab:Sim} for the CelebA dataset. See examples of generated images in Fig.~\ref{fig:Examples_CelebA}. In CelebA, we do not have access to ground-truths, so we perform biometric matching with disjoint samples of the subject not used in the training set. We refer this as the `simulation' result. We achieve the best biometric matching using the initial loss settings of latent diffusion (Eqn.~\ref{Eqn:SD}). The biometric matching impacts the similarity between generated and gallery images and does not quantify the success of age editing. On the other hand, the generated images using contrastive loss\footnote{We also compare with VGGFace-based biometric loss (Eqn.~\ref{Eqn:SD_biom}), and observed contrastive loss outperforms biometric loss. See Sec.~\ref{custom}.} (Eqn.~\ref{Eqn:SD_contrast}) successfully accomplish aging/de-aging but achieve low matching as the ArcFace model is not trained on generated images. 

Therefore, we conduct an additional experiment of fine-tuning the ArcFace model on subject disjoint age-edited images ($\sim$3,400) and then repeat the matching experiments for the CelebA dataset. We report the \textbf{original-original}, \textbf{modified-modified}, \textbf{original-modified} (before fine-tuning ArcFace) and \textbf{modified-modified} (after fine-tuning ArcFace) face matching performance and the corresponding DET curves in Fig.~\ref{Fig:CelebA_Mod} for the contrastive loss and \textit{wzx} token combination. Note that there is a significant improvement in face matching performance between the modified-modified images and original-modified images after fine-tuning. We achieve \textbf{FNMR=3\% at FMR=0.01\%} and \textbf{FNMR=1\% at FMR = 0.1\%} with the fine-tuned face matcher on the age-edited images. We down-sample the modified images to the same resolution as the original images, and observe similar results. Additionally, the fine-tuned face matcher drastically improves when comparing original-modified images, indicating that synthetic images can improve the robustness of existing face matchers as suggested in~\cite{Synface}. 

We report the biometric matching performance using the ArcFace matcher between \textbf{original and modified} images for the Age DB dataset. In AgeDB, we have a separate gallery set consisting of images across age groups different than the images used during training. We use them as ground-truth for evaluation and refer this as the `imputation' result. As anticipated, we observe modest performance across a majority of the age groups barring `child'. We had only 28 images from 18 subjects (out of 100) corresponding to child group, and some of the images were of extremely poor quality, thereby resulting in an abnormal high value of FNMR. See examples of generated images in Fig.~\ref{fig:Examples_ageDB}. We present the the DET curves and the corresponding FNMR values @FMR=0.01/0.1\% in Fig.~\ref{fig:agedbDET}. 

\begin{figure}[ht!]
    \centering
         \includegraphics[width=0.4\textwidth]{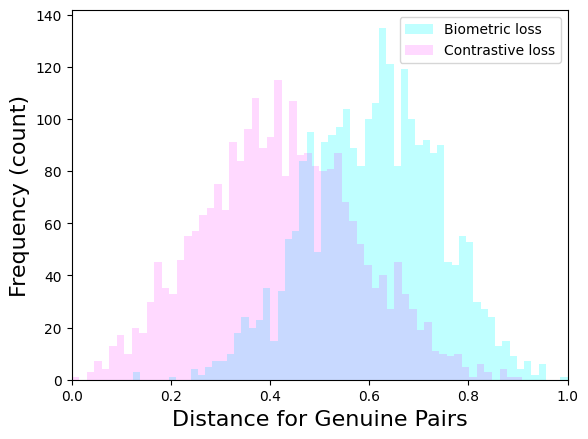}
    \caption{Comparison of auxiliary loss functions (VGGFace-based biometric loss vs. Contrastive loss) in terms of cosine distance scores computed for genuine pairs using the ArcFace matcher. Contrastive loss produces desirable lower distance between genuine pairs.}
    \label{fig:Biomcontrast}
\end{figure}

\subsection{Comparison of auxiliary loss functions}
\label{custom}
We compare the proposed loss functions: 1) VGGFace-based Biometric loss and 2) Contrastive loss and observe a reduction in FNMR up to 46\% @FMR=0.01\% averaged across all age groups when using contrastive loss with respect to biometric loss. Genuine match scores (scores between original and age-edited images of the same individual) that indicate intra-class fidelity are much better preserved when using contrastive loss (see Fig.~\ref{fig:Biomcontrast}). We explored different values of $\lambda_b$ and $\lambda_s$, $=\{0.01, 0.1, 1, 10\}$, and observe 0.1 produces the best results for both variables.


\begin{figure}[ht!]
\centering
   \begin{subfigure}[b]{0.42\textwidth}   
         \includegraphics[width=\textwidth]{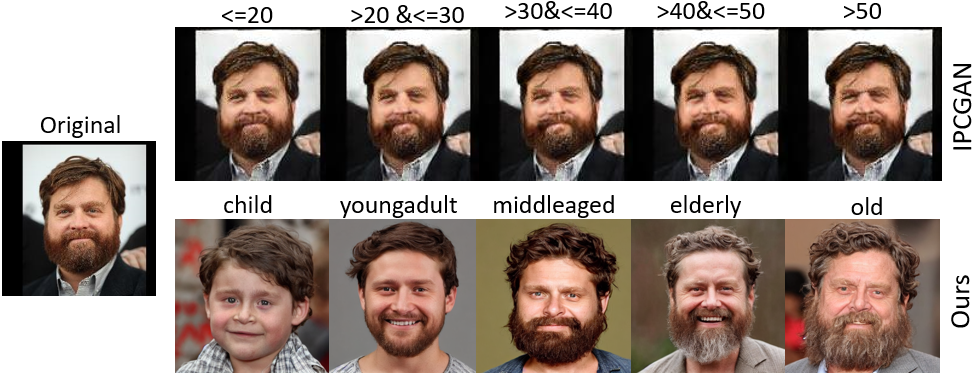}
        \end{subfigure} 
     \begin{subfigure}[b]{0.42\textwidth}  
         \includegraphics[width=\textwidth]{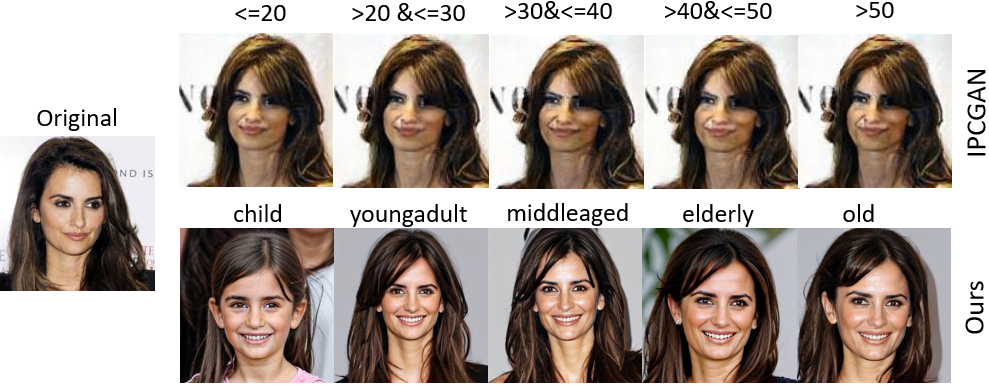}
        \end{subfigure} \vspace{0.5em}
    \begin{subfigure}{0.42\textwidth}
    \hspace{1.5em}
         \includegraphics[width=0.82\textwidth]{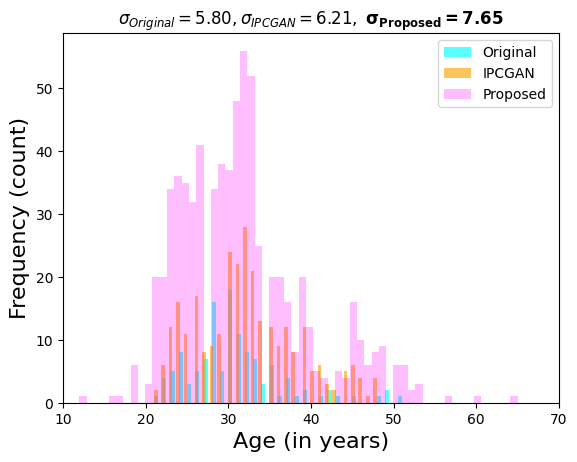}
    \end{subfigure}
    \caption{(Top and middle): Comparison of outputs produced by IPCGAN and the proposed method. (Bottom): Age predictions by automated age predictor shows that our method generates images with a wider age dispersion compared to original CACD images and IPCGAN-generated images.}
    \label{fig:IPCGAN}
\end{figure}

\begin{figure}[ht!]
     \centering
     \begin{subfigure}[b]{0.06\textwidth}
         \centering
         \caption*{\textbf{original}}
         \includegraphics[width=\textwidth, height=\textwidth]{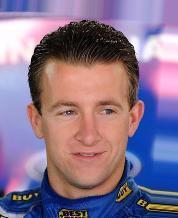}      
     \end{subfigure} 
     \begin{subfigure}[b]{0.06\textwidth}
         \centering
         \caption*{AttGAN}
         \includegraphics[width=\textwidth]{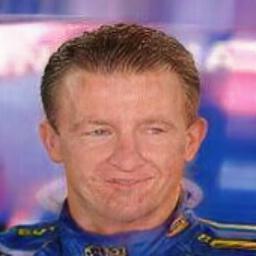}     
     \end{subfigure}
     \begin{subfigure}[b]{0.06\textwidth}
         \centering
          \caption*{Talk-to-Edit}
         \includegraphics[width=\textwidth]{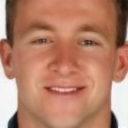}
     \end{subfigure}
     \begin{subfigure}[b]{0.06\textwidth}
         \centering
          \caption*{Proposed}
         \includegraphics[width=\textwidth]{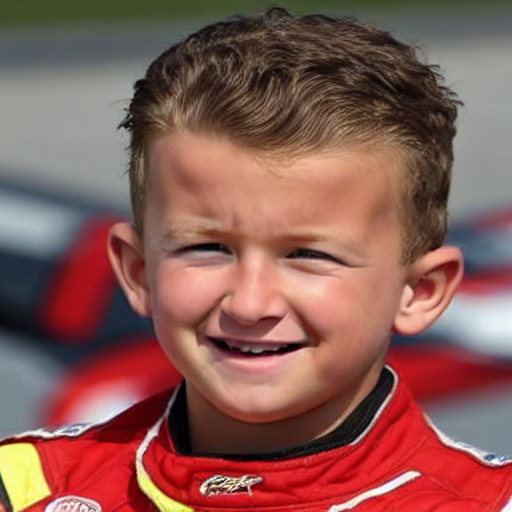}
     \end{subfigure}
     \begin{subfigure}[b]{0.06\textwidth}
         \centering
          \caption*{AttGAN}
         \includegraphics[width=\textwidth]{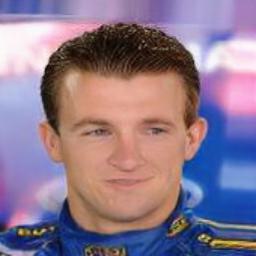}
     \end{subfigure}
     \begin{subfigure}[b]{0.06\textwidth}
         \centering
          \caption*{Talk-to-Edit}
         \includegraphics[width=\textwidth]{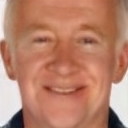}
     \end{subfigure}
     \begin{subfigure}[b]{0.06\textwidth}
         \centering
          \caption*{Proposed}
         \includegraphics[width=\textwidth]{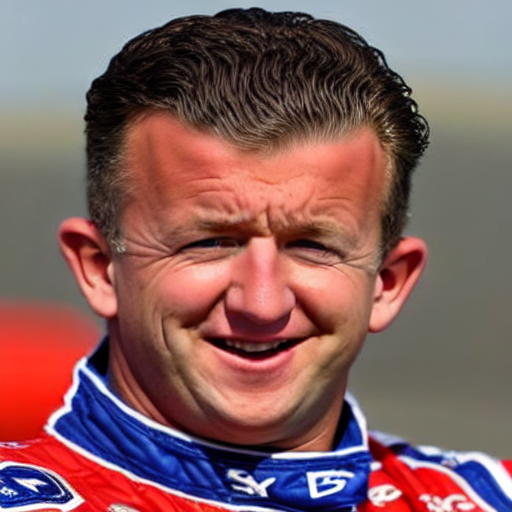}
     \end{subfigure} \\
     \begin{subfigure}[b]{0.06\textwidth}
         \centering
         \includegraphics[width=\textwidth, height=\textwidth]{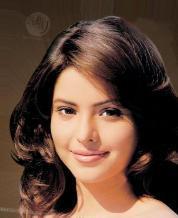}
     \end{subfigure} 
     \begin{subfigure}[b]{0.06\textwidth}
         \centering
         \includegraphics[width=\textwidth]{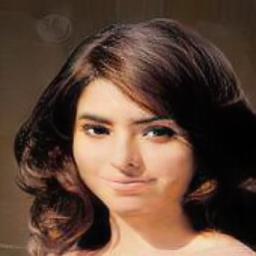}
     \end{subfigure}
     \begin{subfigure}[b]{0.06\textwidth}
         \centering
         \includegraphics[width=\textwidth]{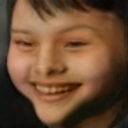}
     \end{subfigure}
     \begin{subfigure}[b]{0.06\textwidth}
         \centering
         \includegraphics[width=\textwidth]{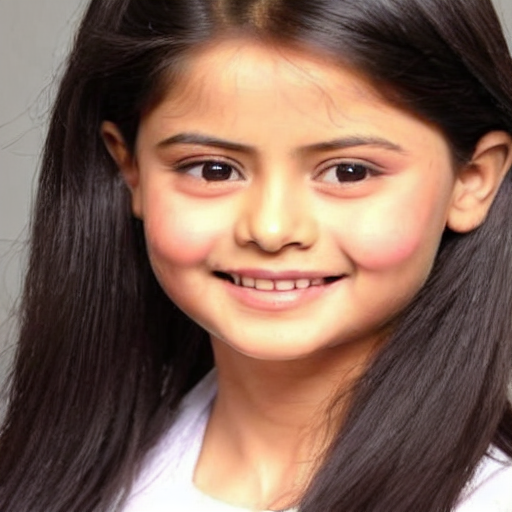}     
     \end{subfigure} 
     \begin{subfigure}[b]{0.06\textwidth}
         \centering
         \includegraphics[width=\textwidth]{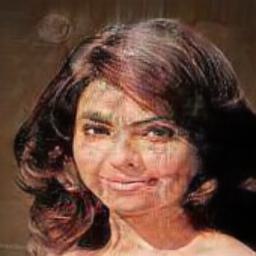}
         \end{subfigure}
     \begin{subfigure}[b]{0.06\textwidth}
         \centering
         \includegraphics[width=\textwidth]{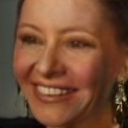}     
     \end{subfigure}
     \begin{subfigure}[b]{0.06\textwidth}
         \centering
         \includegraphics[width=\textwidth]{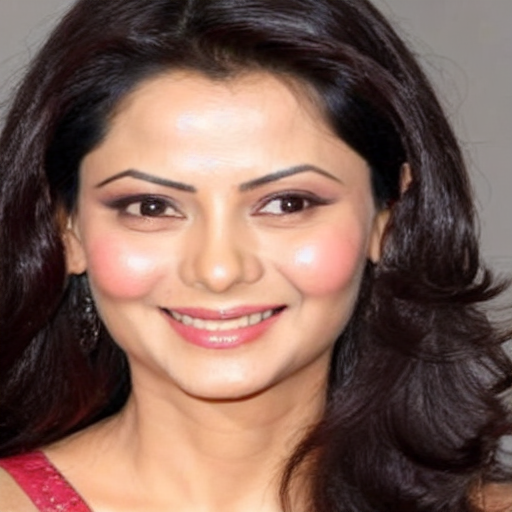}     
     \end{subfigure}\vspace{0.5em}  
    \begin{subfigure}[t]{0.4\textwidth}
    \hspace{-1.0em} 
    \begin{tabular}{l||lll} \hline
     \multirow{2}{*}{Age group} & \multicolumn{3}{c}{Methods}         \\ \cline{2-4}
                           & AttGAN    & Talk-to-Edit & Proposed \\ \hline \hline
child                      & -         &   0.99/0.40           &\textbf{0.56/0.26}           \\
teenager                   & -         &   1.0/0.50           & \textbf{0.29/0.10}         \\
youngadults                & 0.47/0.20 & 0.70/0.21    &  \textbf{0.28/0.08 }      \\
middleaged                 & -         &  0.51/0.13          & \textbf{0.27/0.09}         \\
elderly                    & -         &  0.83/0.39          & \textbf{0.25/0.09}      \\
old                        & 0.31/0.11 & 0.56/0.22          & \textbf{0.29/0.11}       \\ \hline \hline
Average                    & 0.39/0.15 & 0.76/0.31          &\textbf{0.32/0.12}
\end{tabular}
    \end{subfigure}
    \vspace{0.5em}
  \caption{(Top): Comparison of `young' outputs (columns 2-4) and `old' outputs (columns 5-7) generated by the proposed method with baselines: AttGAN and Talk-to-Edit. The original images are in the first column. (Bottom): False Non-Match Rate (FNMR) at False Match Rate (FMR) = 0.01/0.1\%}
    \label{fig:Baseline_Young}
\end{figure}

\begin{figure}
\centering
\begin{subfigure}[b]{0.1\textwidth}
         \centering
         \includegraphics[width=\textwidth]{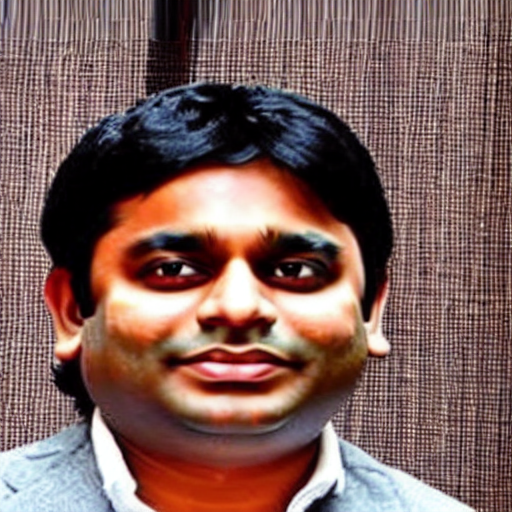}
         \caption{\textit{sks}}
     \end{subfigure}
      \begin{subfigure}[b]{0.1\textwidth}
         \centering
         \includegraphics[width=\textwidth]{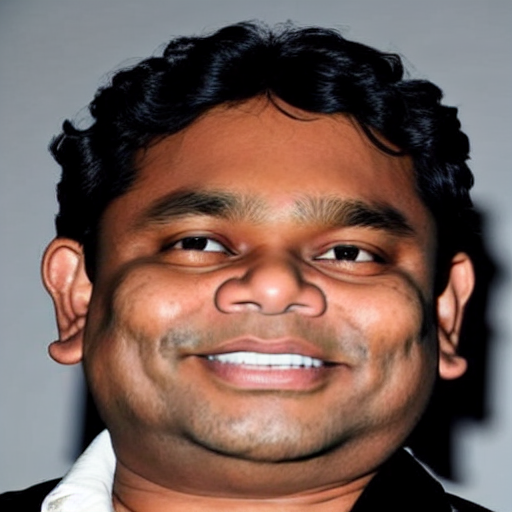}
         \caption{\textit{ukj}}
     \end{subfigure}
     \begin{subfigure}[b]{0.1\textwidth}
         \centering
         \includegraphics[width=\textwidth]{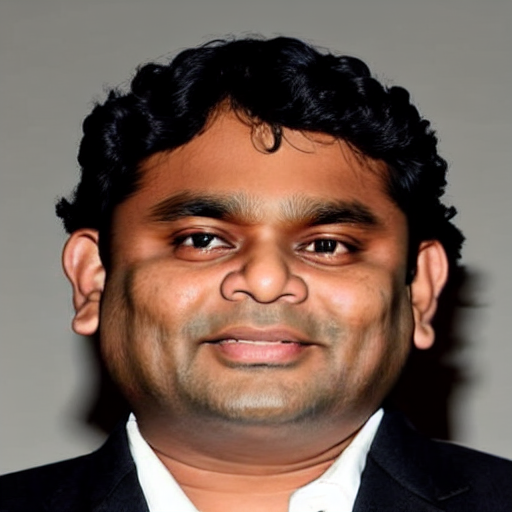}
         \caption{\textit{wzx}}
     \end{subfigure}
     \begin{subfigure}[b]{0.1\textwidth}
         \centering
         \includegraphics[width=\textwidth]{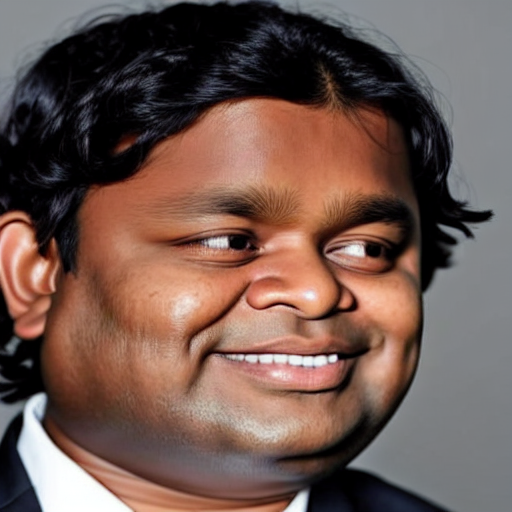}
         \caption{\textit{ams}}
     \end{subfigure}
        \caption{Comparison of the images generated using  the four tokens in this work.}
        \label{fig:Tokens}
\end{figure}

\begin{figure*}[t]
     \centering
     \begin{subfigure}[b]{0.10\textwidth}
         \centering
         \caption*{\textbf{original}}
         \includegraphics[width=\textwidth, height=\textwidth]{images/SubID_0001_NumImages_29_000023.jpg}
        \end{subfigure} \hfill
     \begin{subfigure}[b]{0.10\textwidth}
         \centering
         \includegraphics[width=\textwidth, height=\textwidth]{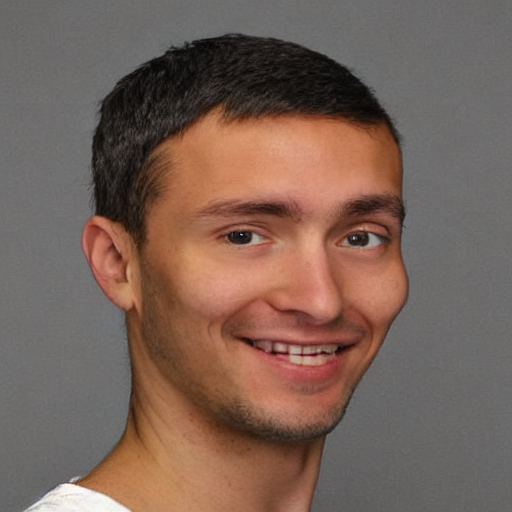}
        \end{subfigure} 
     \begin{subfigure}[b]{0.10\textwidth}
         \centering
         \includegraphics[width=\textwidth]{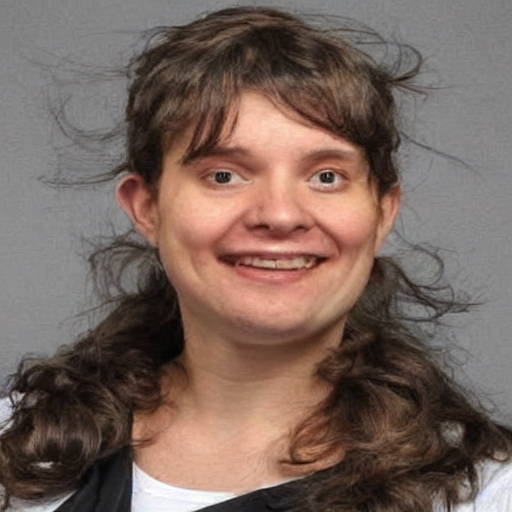}  
     \end{subfigure} 
     \begin{subfigure}[b]{0.10\textwidth}
         \centering
         \includegraphics[width=\textwidth, height=\textwidth]{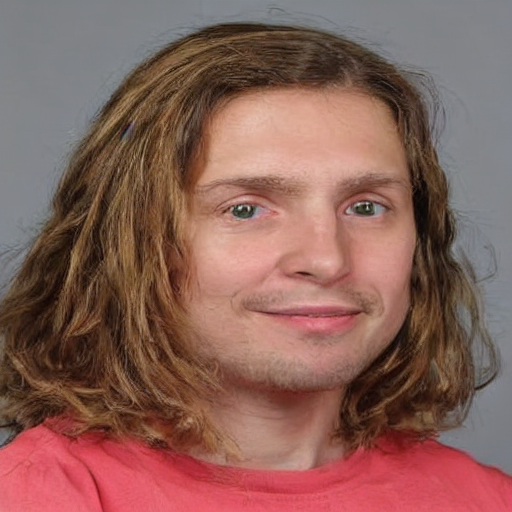}
        \end{subfigure} 
     \begin{subfigure}[b]{0.10\textwidth}
         \centering
         \includegraphics[width=\textwidth]{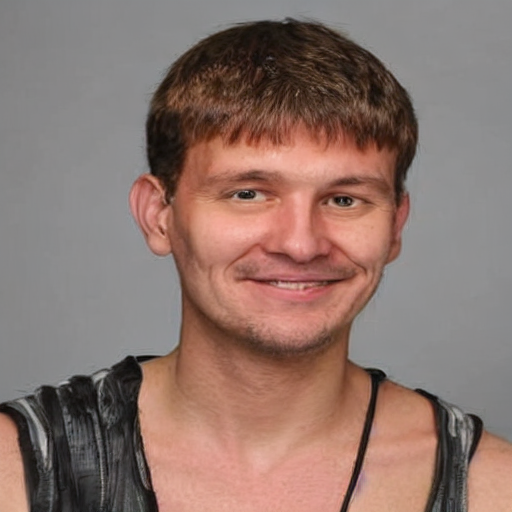}
     \end{subfigure} 
     \begin{subfigure}[b]{0.10\textwidth}
         \centering
         \includegraphics[width=\textwidth, height=\textwidth]{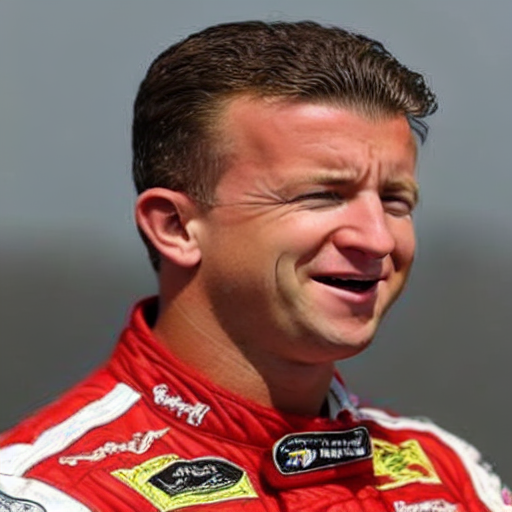}
        \end{subfigure} 
     \begin{subfigure}[b]{0.10\textwidth}
         \centering
         \includegraphics[width=\textwidth]{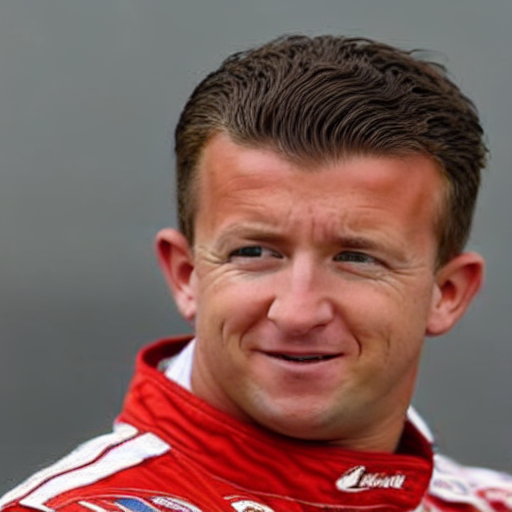}
     \end{subfigure}
     \begin{subfigure}[b]{0.10\textwidth}
         \centering
         \includegraphics[width=\textwidth, height=\textwidth]{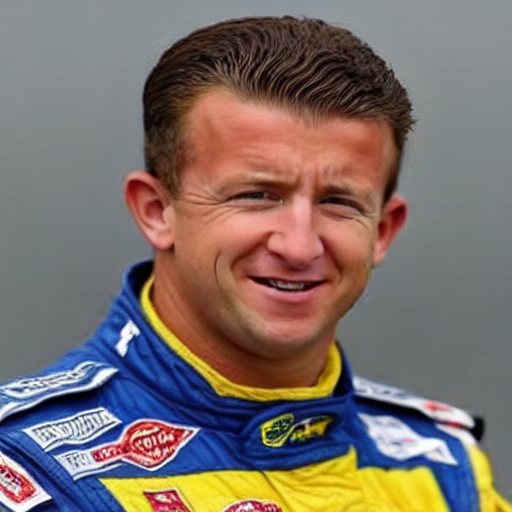}
        \end{subfigure}
     \begin{subfigure}[b]{0.10\textwidth}
         \centering
         \includegraphics[width=\textwidth]{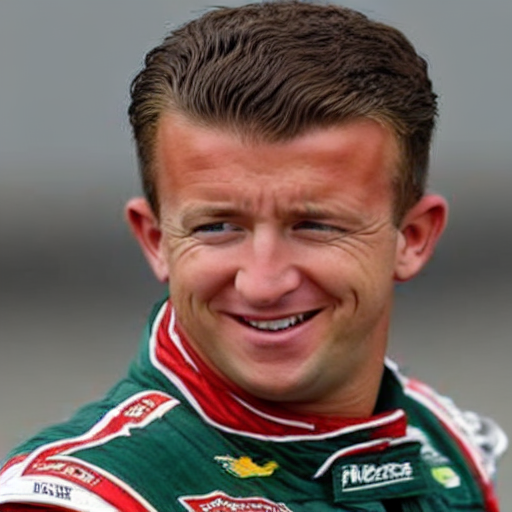}
        
     \end{subfigure}
     \caption{Impact of token (\textit{wzx}) and class label (\textit{person}) on generated images: ``photo of a person" (left) vs. ``photo of a \textbf{\textit{wzx}} person" (right). Note the token is strongly associated with a specific identity belonging to that class.}
        \label{fig:tokenclass}
\end{figure*}

\subsection{Comparison with existing methods}
\label{baseline}
We use IPCGAN~\cite{IPCGAN}, AttGAN~\cite{AttGAN} and Talk-to-Edit~\cite{talktoedit} as baselines for comparison. We evaluate using the pre-trained models of the baselines provided by the authors. As IPCGAN was trained on the CACD dataset~\cite{CACD}, we fine-tune our method on 62 subjects from the CACD dataset. We observe an FNMR=2\% (IPCGAN), compared to FNMR=11\% (Ours) @ FMR=0.01. IPCGAN defaults to the original when it fails to perform aging or de-aging resulting in spuriously low FNMR. We perform automated age prediction using the DeepFace~\cite{DeepFace_Github} age predictor. We observe the images synthesized by our method result in wider dispersion of age predictions compared to the original images and the IPCGAN-generated images, indicating successful age editing. See Fig.~\ref{fig:IPCGAN}. We apply AttGAN and Talk-to-Edit on the CelebA dataset. See comparison between generated images of proposed and baseline methods, and biometric matching performance in Fig.~\ref{fig:Baseline_Young}. We observe that the proposed method (contrastive loss, \textit{sks}) outperforms AttGAN by 19\% on `young' images and by 7\% on `old' images at FMR=0.01. AttGAN can only edit to young or old ages. Further, we observe that the method outperforms Talk-to-Edit by an average FNMR =44\% at FMR=0.01. The different age groups are simulated using a target value parameter in Talk-to-Edit that varies from 0 to 5, each value representing an age group. However, we observe several cases of distorted or absence of outputs in Talk-to-Edit.

\begin{figure}[h]
     \centering
     
         \includegraphics[width=0.45\textwidth]{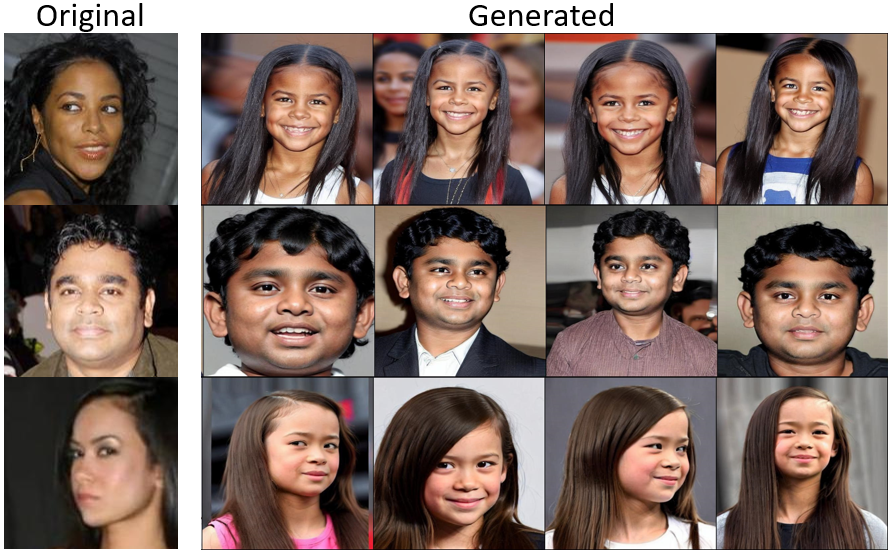}
         
        \caption{Examples of generated images pertaining to diverse sex and ethnicity for `child' group.}
        \label{fig:Ethnic}
\end{figure}
\begin{figure}[h]
     \centering
     \begin{subfigure}[b]{0.12\textwidth}
         \centering
         \includegraphics[width=\textwidth]{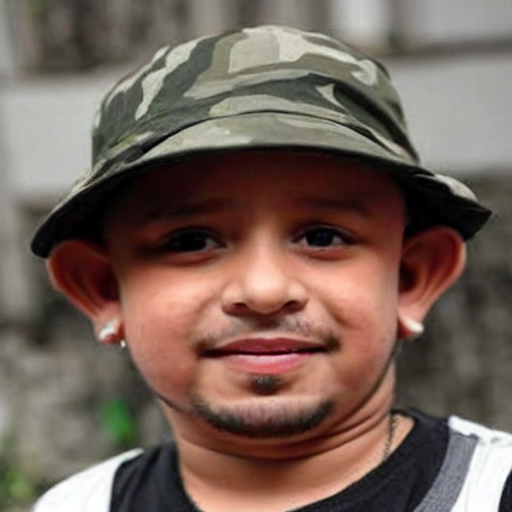} 
     \end{subfigure} 
     \begin{subfigure}[b]{0.12\textwidth}
         \centering
         \includegraphics[width=\textwidth]{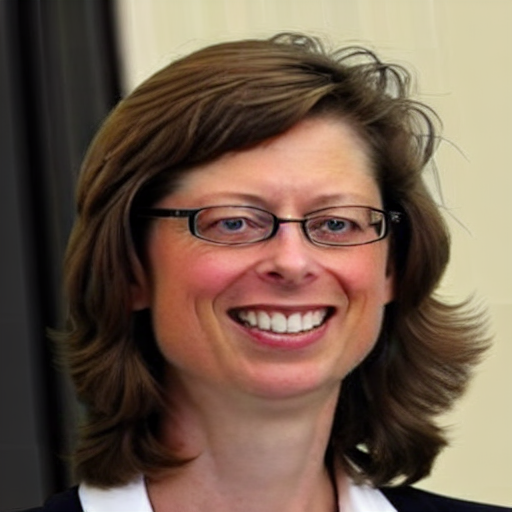}     
     \end{subfigure} 
      \begin{subfigure}[b]{0.12\textwidth}
         \centering
         \includegraphics[width=\textwidth]{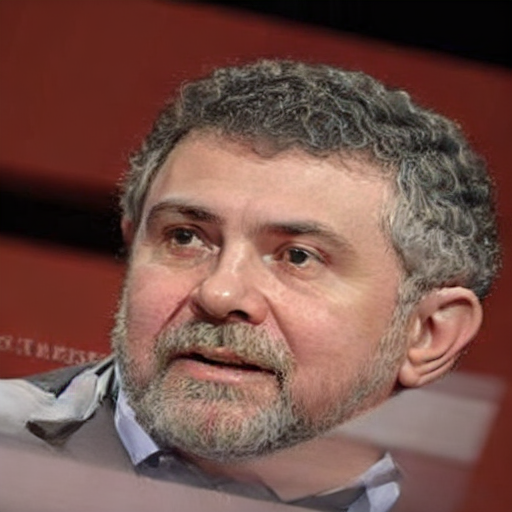}     
     \end{subfigure} 
        \caption{Failure cases corresponding to `child' age group.}
        \label{fig:fail}
\end{figure}


\subsection{User study}
We collected 26 responses from the user study. Rank-1 biometric identification accuracy (averaged across the total number of responses) is equal to 78.8\%. The correct identification accuracy of the age groups are: child = 99.6\%, teenager = 72.7\%, youngadults = 68.1\%, middleaged = 70.7\% and elderly = 93.8\%. The users were able to successfully distinguish between generated images from different age groups with reasonably high accuracy. 

\subsection{Effect of rare tokens}
\label{token}
We use four tokens in this work, namely, \{\textit{sks}, \textit{ukj}, \textit{ams}, \textit{wzx}\}, for the sake of brevity. We observe \textit{sks} and \textit{wzx} tokens result in visually compelling results compared to the remaining two tokens, and have been used for further evaluation. Note these tokens are condensed representations provided by the tokenizer that are determined by identifying rare phrases in the vocabulary (see Fig.~\ref{fig:Tokens}). Additionally, we evaluate the effect of the token and the class label in the prompt in Fig.~\ref{fig:tokenclass}; removing the token results in lapse in identity-specific features. 
\subsection{Effect of demographics}
\label{sec:fail}
We also observed the following effects. \textbf{Age:} The generated images can capture different age groups well if the training set contains images in the middle-aged category. We observe that if training set images comprise mostly elderly images, then the method struggles to render images in the other end of the spectrum, \textit{i.e.}, the child category, and vice-versa. We also observe that we obtain visually compelling results of advanced aging when we use `elderly' in the prompt instead of `old'. 
\textbf{Sex:} The generated images can effectively translate the training images into older age groups for men compared to women. This can be due to the use of makeup in the training images. \textbf{Ethnicity:} We do not observe any strong effects of ethnicity/race variations in the outputs. See Fig.~\ref{fig:Ethnic}. 
Although in some cases, the proposed method struggles with generating `child' images if most of the training images belong to elderly people or contain facial hair. See Fig.~\ref{fig:fail}.

\section{Conclusion}
\label{sec:sum}
Existing facial age editing methods typically struggle with identity-preserved age translation. In this work, we harness latent diffusion coupled with biometric and contrastive losses for enforcing identity preservation while performing facial aging and de-aging. We use a regularization image set to impart the understanding of age progression and regression to the diffusion model, that in turn, transfers the effects onto an unseen individual while preserving their identity. The generation process is guided by intuitive text prompts indicating the desired age. Our method demonstrates significantly better results in terms of both qualitative and quantitative evaluation, and outperforms existing methods with a reduction in FNMR up to 44\% at FMR=0.01\%.

Future work will focus on designing zero-shot age editing without fine-tuning, and utilizing composable diffusion models~\cite{composablediff} for fine-grained age editing.

\clearpage
\balance
{\small
\bibliographystyle{ieee}
\bibliography{egbib}
}

\end{document}